\documentclass[runningheads]{llncs}
\usepackage{graphicx}

\usepackage{caption}
\usepackage{microtype}
\usepackage{booktabs}
\usepackage{multirow} 
\usepackage{marvosym}
\usepackage{cite}
\usepackage{multirow}
\usepackage[section]{placeins}
\usepackage{todonotes}

\newcommand{\refdef}[1]{Definition~\ref{#1}}

\newcommand{\refeq}[1]{Eq.~\eqref{#1}}

\usepackage{listings}
\usepackage{ifxetex}
\usepackage{fancyhdr}
\usepackage{hyperref}
\usepackage{graphicx}
\usepackage{wrapfig}
\usepackage{amsmath}
\usepackage{amsfonts}
\usepackage{amssymb}
\usepackage{listings}
\usepackage{setspace}
\usepackage{wrapfig}
\usepackage{tablefootnote}
\usepackage{multirow}
\usepackage{booktabs}
\usepackage[section]{placeins}
\usepackage{color}
\usepackage{amsfonts}
\usepackage{ulem}
\usepackage{algorithm}
\usepackage{algorithmicx}
\usepackage{algpseudocode}
\usepackage{graphicx}
\usepackage{float}
\usepackage{dsfont}
\usepackage{cleveref}

\newcommand{\mat}[1]{\mathbf{#1}}
\newcommand{\set}[1]{\mathcal{#1}}
\newcommand{\pnorm}[1]{\lVert{#1}\rVert}

\DeclareMathOperator*{\loss}{{\ell}}
\DeclareMathOperator*{\dist}{{d}}

\DeclareMathOperator*{\sgd}{sgd}

\DeclareMathOperator*{\SetSymMat}{\mathcal{S}}

\DeclareMathOperator*{\argmin}{\arg\min}

\newcommand{\RN}{\mathbb{R}}

\newcommand{\lOne}{l_1}
\newcommand{\lZero}{l_0}

\DeclareMathOperator*{\regularization}{\ensuremath{{\theta}}}
\newcommand{\x}{\ensuremath{\vec{x}}}

\newcommand{\y}{\ensuremath{y}}

\newcommand{\xorig}{\ensuremath{\vec{x}_{\text{orig}}}}

\newcommand{\setX}{\ensuremath{\set{X}}}
\newcommand{\setY}{\ensuremath{\set{Y}}}

\newcommand{\xcf}{\ensuremath{\vec{x}_{\text{cf}}}}

\newcommand{\ycf}{\ensuremath{y'}}

\newcommand{\q}{\ensuremath{\vec{q}}}
\newcommand{\dimsym}{d}

\newcommand{\classifier}{\ensuremath{h}}

\newcommand{\protolabel}{\ensuremath{o}}
\DeclareMathOperator*{\prototype}{{p}}
\DeclareMathOperator*{\distmat}{{\mat{\Omega}}}

\DeclareMathOperator*{\prototypedist}{\dist}

\newcommand{\reject}{\ensuremath{r}}
\newcommand{\threshold}{\ensuremath{\theta}}
\newcommand{\relsim}{\ensuremath{r_\text{RelSim}}}
\newcommand{\rel}{\ensuremath{r}}
\newcommand{\rdist}{\ensuremath{r_\text{Dist}}}
\newcommand{\rprob}{\ensuremath{r_\text{Proba}}}

\usepackage{xspace}
\newcommand{\ie}{i.e.\@\xspace}
\newcommand{\eg}{e.g.\@\xspace}

\newcommand{\proto}{\vec{\prototype}}
\newcommand{\protop}{\vec{\prototype}_{+}}
\newcommand{\protom}{\vec{\prototype}_{-}}
\newcommand{\transp}[1]{#1^\top}
\newcommand{\prob}{p}

\newcommand{\rej}{R}

\begin{document}
\title{Explaining Reject Options of\\ Learning Vector Quantization Classifiers
\thanks{We gratefully acknowledge funding from
the Deutsche Forschungsgemeinschaft (DFG, German Research
Foundation) for grant TRR 318/1 2021 – 438445824, and the VW-Foundation for the project \textit{IMPACT} funded in the frame of the funding line \textit{AI and its Implications for Future Society}.}}

\titlerunning{Explaining Reject Options of LVQ}

\author{Andr\'e Artelt\inst{1}, Johannes Brinkrolf\inst{1}, Roel Visser\inst{1} \and Barbara Hammer\inst{1}}
\authorrunning{A. Artelt et al.}
\institute{CITEC - Cognitive Interaction Technology\\Bielefeld University, 33619 Bielefeld, Germany\\
\email{\{aartelt,jbrinkro,rvisser,bhammer\}@techfak.uni-bielefeld.de}}
\maketitle              % typeset the header of the contribution
\begin{abstract}
%The abstract should briefly summarize the contents of the paper in 15--250 words.
While machine learning models are usually assumed to always output a prediction, there also exist extensions in the form of reject options which allow the model to reject inputs where only a prediction with an unacceptably low certainty would be possible. With the ongoing rise of eXplainable AI, a lot of methods for explaining model predictions have been developed. However, understanding why a given input was rejected, instead of being classified by the model, is also of interest. Surprisingly, explanations of rejects have not been considered so far.

We propose to use counterfactual explanations for explaining rejects and investigate how to efficiently compute counterfactual explanations of different reject options for an important class of models, namely prototype-based classifiers such as learning vector quantization models.
\end{abstract}
\section{Introduction}
Nowadays, machine learning (ML) based decision making systems are increasingly used in safety-critical or high-impact applications like autonomous driving~\cite{AutonmousDriving}, credit (risk) assessment~\cite{CreditRiskML} and predictive policing~\cite{PredictivePolicing}. Because of this, there is an increasing demand for transparency which was also recognized by the policy makers and emphasized in legal regulations like the EU's GDPR~\cite{gdpr}. It is a common approach to realize transparency by explanations -- i.e. providing an explanation of why the system behaved in the way it did -- which gave rise to the field of explainable artificial intelligence (XAI or eXplainable AI)~\cite{SurveyXai, ExplainableArtificialIntelligence}. Although it is still unclear what exactly makes up a ``good'' explanation~\cite{doshivelez2017rigorous,offert2017i}, a lot of different explanation methods have been developed~\cite{ExplainingBlackboxModelsSurvey,molnar2019}. Popular explanations methods~\cite{molnar2019,SurveyXai} are feature relevance/importance methods~\cite{FeatureImportance} and examples based methods~\cite{CaseBasedReasoning}. Instances of example based methods are contrasting explanations like counterfactual explanations~\cite{CounterfactualWachter,CounterfactualReviewChallenges} and prototypes \& criticisms~\cite{PrototypesCriticism} -- these methods use a set or a single example for explaining the behavior of the system.

Another important aspect, in particular in safety-critical and high-risk applications, is reliability: If the system is not ``absolutely'' certain about its decision/prediction it might be better to refuse making a prediction and, for instance, pass the input back to human -- if making mistakes is ``expensive'' or critical, the system should reject inputs where it is not certain enough in its prediction. As a consequence, a mechanism called reject options has been pioneered~\cite{ChowReject}, where optimal reject rates are determined based on costs assigned to miss-classifications (e.g. false positives and false negatives) and rejects. Many realizations of reject options are based on probabilities, like class probabilities in classification~\cite{ChowReject}. However, not all models output probabilities along with their predictions or if they do, their computed probabilities might be of ``bad quality'' (e.g. just a score in $[0,1]$ without any probabilistic/statistical foundation). One possible remedy is to use a general (i.e. model agnostic) post-processing method or wrapper on top of the model for computing reliable certainty scores, as is done in~\cite{ClassificationRejectOptionHerbei2006}, or use a method like conformal prediction~\cite{ConformalShaferV08} in which a non-conformity measure together with a hold-out data set is used for computing certainties and confidences of predictions. Another option is to develop model specific certainty measures and consequently reject options, which for instance was done for prototype-based methods like learning vector quantization (LVQ) models. These are a class of models that convinced with simplicity (and thus interpretability), while still being powerful predictive models, and have also excelled in settings like life-long learning and biomedical applications~\cite{LvqReview, LifeLongLvq, OnlineIncrementalLvq, LvqPlantDiseasesSpectralDataOwomugishaNQBM20, RejectProbaLvqBrinkrolfH20,OptimumRejectLvqFischerHW15,EfficientRejectLvqFischerHW15}.

We think that while explaining the prediction of the model is important, similarly explaining why a given input was rejected is also important -- i.e. explaining why the model ``thinks'' that it does not know enough for making a proper and reliable prediction.
\textit{For instance, consider a biomedical application where a model is supposed to assist in some kind of early cancer detection -- such a scenario could be considered a high-risk application where the addition of a reject option to the model would be necessary. In such a scenario, it would be very useful if a reject is accompanied with an explanation of why this sample was rejected, because by this we could learn more about the model and the domain itself -- e.g. an explanation could state that some combination of serum values is very unusual compared to what the model has seen before.}
Surprisingly, (to the best of our knowledge) explaining rejects have not been considered so far.

\paragraph{Contributions}
In this work, we focus on prototype-based models such as learning vector quantization (LVQ) models and propose to explain LVQ reject options by means of counterfactual explanations -- i.e. explaining why a particular sample was rejected by the LVQ model. Whereby we consider different reject options separately and study how to efficiently compute counterfactual explanations in each of these cases.

The remainder of this work is structured as follows: After reviewing the foundations of learning vector quantization (Section~\ref{sec:foundations:lvq}), popular reject options for LVQ (Section~\ref{sec:foundations:rejectoptions}), and counterfactual explanations (Section~\ref{sec:foundations:counterfactuals}), we propose a modeling and algorithms for efficiently computing counterfactual explanations of different LVQ reject options (Section~\ref{sec:modeling}). In Section~\ref{sec:experiments}, we empirically evaluate our proposed modelings and algorithms on several different data sets with respect to different aspects. Finally, this work closes with a summary and conclusion in Section~\ref{sec:conclusion}.
Note that for the purpose of readability, all proofs and derivations are given in Appendix~\ref{sec:appendix}.

\section{Foundations}\label{sec:foundations}
In the following, we briefly review the foundations our work is based on. 

\subsection{Learning Vector Quantization}\label{sec:foundations:lvq}
In this work, we focus on learning vector quantization (LVQ) models. In modern variants, these models constitute state-of-the-art classifiers which can also be used for incremental, online, and federated learning~\cite{biehl_prototype-based_2016,DBLP:conf/esann/GepperthH16,losing_incremental_2018,Brinkrolf2021ESANN}.
In its basic variant as proposed in~\cite{kohonen_self-organizing_1997}, it relies on Hebbian learning. However, further extensions based on cost functions have been introduced -- these have the benefit that they can be easily extended to a more flexible metric learning scheme~\cite{schneider_adaptive_2009}. First, we introduce generalized LVQ (GLVQ)~\cite{sato_generalized_1995} and then take a look at extensions for metric learning.
In the following, we assume that our the domain is the real valued vector space -- \ie the classification scenario takes place in $ \RN^\dimsym  $ with a fix number of classes $ k $ which are enumerated as $ \{1,\dots,k\}=\setY $.
An LVQ model is characterized by $ m $ labeled prototypes $ (\proto_j,c(\proto_j))\in\RN^\dimsym\times\setY $, $ j\in\{1,\dots,m\} $, whereby the labels $ c(\proto_j) $ of the prototypes are fixed. Classification of a sample $ \x\in\RN^\dimsym $ takes place by a winner takes all %\todo{depending on format in text ref to equation? J: Idk, In the paper, I would write it inline to save some space.}
scheme:
\begin{equation}
	\x \mapsto c(\proto_j)\ \text{ with } j=\argmin_{j\,\in\,\{1,\dots,m\}} \prototypedist(\x,\proto_j)
\end{equation}
where the squared Euclidean distance is used:
\begin{equation}
	\label{eq:squaredEuclidean}
	\prototypedist(\x,\proto_j)=\transp{(\x-\proto_j)}(\x-\proto_j)
\end{equation}
Note that the prototypes' positions not only allow an interpretable classification but also act as a representation of the data and its underlying classes.
Training of LVQ models is done in a supervised fashion -- i.e. based on given labeled samples $ (\x_i,\y_i)\in\RN^\dimsym\times\setY$ with $ i\in\{1,\dots,N\} $. For GLVQ~\cite{sato_generalized_1995}, the learning rule originates from minimizing the following cost function:
\begin{equation}
	\label{eq:lvqcostfct}
	E=\sum_{i=1}^N \sgd\left(\dfrac{\prototypedist(\x_i,\protop)-\prototypedist(\x_i,\protom)}{\prototypedist(\x_i,\protop)+\prototypedist(\x_i,\protom)}\right)
\end{equation}
where $ \sgd(\cdot) $ is a monotonously increasing function -- \eg the logistic function or the identity %\todo{J: which one is used later in the experiments?}.
Furthermore, $ \protop $ denotes the closest prototype with the same label -- \ie $ c(\x)=c(\protop) $ --, and $ \protom $ denotes the closest prototype which belongs to a different class.
Note that the models' performance heavily depends on the suitability of the Euclidean metric for the given classification problem -- note that this is often not the case, in particular in case of different feature relevances.
Because of this, a powerful metric learning scheme has been proposed in~\cite{schneider_adaptive_2009} which substitutes the squared Euclidean metric in~\refeq{eq:squaredEuclidean} by a weighted alternative:
\begin{equation}
	\prototypedist(\x, \proto) = \transp{(\x-\proto)}\distmat(\x-\proto)
\end{equation}
where $ \distmat\in\SetSymMat_{+}^\dimsym $ refers to a positive semidefinite matrix encoding the relevance of each feature. This matrix is treated as an additional parameter which is, together with the prototypes' positions, chosen such that the cost function~\refeq{eq:lvqcostfct} is minimized -- usually, a gradient based method like LBFGS is used. Due to the parameterization of the metric, this LVQ variant is often named generalized matrix LVQ (GMLVQ). Further details can be found in~\cite{schneider_adaptive_2009}.

\subsection{Reject Options}\label{sec:foundations:rejectoptions}
LVQ schemes provide a classification rule which assigns a label to every possible input no matter how reliable \& reasonable such classifications might be. Reject options allow the classifier to reject a sample if a certain prediction is not possible -- \ie the sample is too close to the decision boundary, or it is very different from the observed training data and therefore a classification based on the learned prototypes would not be reasonable.
For LVQ models, as reviewed in the previous section, there exist some reject options which allows the model to deny a classification of a sample if a reliable \& certain classification is not possible. Generally speaking, we extend the set of possible predictions by a new class which represents a reject -- \ie for a classifier $ \classifier:\setX\to\setY $, we construct $ \classifier: \setX\to\setY\cup\{\rej\} $.
Quite a few methods for doing so have been proposed and evaluated in \cite{EfficientRejectLvqFischerHW15}. Those methods use a function $\reject:\setX\to\RN$ for computing the certainty of a prediction by the classifier~\cite{ChowReject}. If the certainty is below some threshold $\threshold$, the sample is rejected, more formally if
\begin{equation}\label{eq:reejctoption}
	\reject(\x) < \threshold
\end{equation}
then $ \classifier(\x)=\rej $.
In the following, we consider three popular realizations of the certainty function $\reject(\cdot)$ in the context of LVQ models.

\paragraph*{Relative similarity:}
In~\cite{EfficientRejectLvqFischerHW15}, a very natural realization of $\reject(\cdot)$ called relative similarity (RelSim) yields excellent results for LVQ models:
\begin{equation}\label{eq:relsim}
	\relsim(\x) = \frac{\prototypedist(\x,\protom) - \prototypedist(\x,\protop)}{\prototypedist(\x,\protom) + \prototypedist(\x,\protop)}
\end{equation}
where $\protop$ denotes the closest prototype and $\protom$ denotes the closest prototype belonging to a different class than $ \protop $. Note that this fraction is always non-negative and smaller than $1$. Obviously, it is $0$ if the distance between the sample and the closest prototype $ \protop $ equals the distance to $ \protom $. At the same time, RelSim gets close to $0$ if the sample is far away -- \ie both distances are large~\cite{Brinkrolf2018AT}.

\paragraph*{Decision boundary distance:}
Another, similar, realization of a certainty measure is the distance to the decision boundary (Dist). In case of LVQ, this can be formalized as follows~\cite{EfficientRejectLvqFischerHW15}:
\begin{equation}\label{eq:rejectdist}
	\rdist(\x) = \frac{|\prototypedist(\x,\protop) - \prototypedist(\x,\protom)|}{2\pnorm{\protop - \protom}_2^2}
\end{equation}
where $\protop$ and $ \protom $ are defined in the same way as in RelSim~\refeq{eq:relsim}. Note that~\refeq{eq:rejectdist} is not normalized and depends on the prototypes and their distances to the sample $ \x $. It is worthwhile mentioning that~\refeq{eq:rejectdist} is closely related to the reject option of a SVM~\cite{platt_probabilistic_1999} -- in case of binary classification problem and a single prototype per class in a LVQ model, both models determine a separating hyperplane.

\paragraph*{Probabilistic certainty:}
The third certainty measure is a probabilistic one. The idea is to obtain proper class probabilities $ \prob(\y\mid\x) $ for each class $ \y\in\setY $ and reject a sample $ \x $ if the probability for the most probable class is lower than a given threshold $ \threshold $. We denote the probabilistic certainty measure as follows:
\begin{equation}\label{eq:probareejct}
	\rprob(\x) = \underset{\y\,\in\setY}\max\;\prob(\y\mid\x).
\end{equation}
In the following, we use the fastest method from~\cite{Brinkrolf2019NCAA} for computing class probabilities given a trained LVQ models. The method~\cite{price_pairwise_1994} combines estimates from binary classifiers in order to obtain class probabilities for the final prediction.
First, because the estimates of a binary classifiers are necessary, we train a single LVQ model for each pair of classes using only the samples belonging to those two classes. This yields a set of $ |\setY|(|\setY|-1)/2 $ binary classifiers. Next, a data-dependent scaling of the predictions follows to yield pairwise probabilities. Here, we are using RelSim~\refeq{eq:relsim} and fit a sigmoid function to the real-valued scores. This mimics the approach by Platt~\cite{platt_probabilistic_1999} which has become very popular in the context of SVMs. This post-processing yields estimates $ \rel_{i,j}(\x) $ of pairwise probabilities for every sample $ \x $ and pairs of classes $ i $ and $ j $:
\begin{equation}
	\rel_{i,j} = \prob(\y=i\mid \y=i \vee j, \x)
\end{equation}
Given those pairwise probabilities and assuming a symmetric completion of the pairs $ \rel_{i,j}+\rel_{j,i}=1 $, the posterior probabilities are obtained as follows:
\begin{equation}\label{eq:pkpd}
	\prob(\y=i\mid\x) = \frac{1}{\sum_{j\neq i}\frac{1}{\rel_{i,j}} - (|\setY| - 2)}
\end{equation}
where $ i,j\in\setY $~\cite{price_pairwise_1994}. After computing all probabilities, a normalization step is required such that $ \sum_{i=1}^k \prob(\y=i\mid\x)=1 $. We refer to~\cite{Brinkrolf2019NCAA} for further details on all these steps.

\subsection{Counterfactual Explanations}\label{sec:foundations:counterfactuals}
Counterfactual explanations (often just called \textit{counterfactuals}) are a prominent instance of contrasting explanations, which state a change to some features of a given input such that the resulting data point, called the counterfactual, causes a different behavior of the system than the original input does. Thus, one can think of a counterfactual explanation as a suggestion of actions that change the model's behavior/prediction. One reason why counterfactual explanations are so popular is that there exists evidence that explanations used by humans are often contrasting in nature~\cite{CounterfactualsHumanReasoning} -- i.e. people often ask questions like \textit{``What would have to be different in order to observe a different outcome?''}.
For illustrative purposes, consider the example of loan application: \textit{Imagine you applied for a credit at a bank. Unfortunately, the bank rejects your application. Now, you would like to know why. In particular, you would like to know what would have to be different so that your application would have been accepted.
A possible explanation might be that you would have been accepted if you had earned 500\$ more per month and if you would not have a second credit card.}
Despite their popularity, the missing uniqueness of counterfactuals could pose a problem: Often there exist more than one possible \& valid counterfactual -- this is called the Rashomon effect~\cite{molnar2019} -- and in such cases, it is not clear which or how many of them should be presented to the user. One common modeling approach (if this problem is not simply ignored) is to enforce uniqueness by a suitable formalization.

In order to keep the explanation (suggested changes) simple -- i.e. easy to understand -- an obvious strategy is to look for a small number of changes so that the resulting sample (counterfactual) is similar/close to the original sample, which is aimed to be captured by~\refdef{def:counterfactual}.
\begin{definition}[(Closest) Counterfactual Explanation~\cite{CounterfactualWachter}]\label{def:counterfactual}
Assume a prediction function $\classifier:\RN^\dimsym \to \setY$ is given. Computing a counterfactual $\xcf \in \RN^\dimsym$ for a given input $\xorig \in \RN^\dimsym$ is phrased as an optimization problem:
\begin{equation}\label{eq:counterfactualoptproblem}
\underset{\xcf \,\in\, \RN^\dimsym}{\arg\min}\; \loss\big(\classifier(\xcf), \ycf\big) + C \cdot \regularization(\xcf, \xorig)
\end{equation}
where $\loss(\cdot)$ denotes a loss function, $\ycf$ the target prediction, $\regularization(\cdot)$ a penalty for dissimilarity of $\xcf$ and $\xorig$, and $C>0$ denotes the regularization strength.
\end{definition}
The counterfactuals from~\refdef{def:counterfactual} are also called \textit{closest counterfactuals} because the optimization problem~\refeq{eq:counterfactualoptproblem} tries to find an explanation $\xcf$ that is as close as possible to the original sample $\xorig$. However, other aspects like plausibility and actionability are ignored in~\refdef{def:counterfactual}, but are covered in other work~\cite{CounterfactualGuidedByPrototypes,PlausibleCounterfactuals,ActionableCounterfactuals}. Note that it is not always clear which type of counterfactual is meant when people talk about counterfactuals -- in this work, we use the term counterfactuals in the spirit of~\refdef{def:counterfactual}.

\section{Counterfactual Explanations of Reject}\label{sec:modeling}
In this section, we elaborate our proposal of using counterfactual explanations for explaining LVQ reject options.
First, we introduce the general modeling in Section~\ref{sec:modeling:modeling}, before then investigating the computational aspects of each reject option in Section~\ref{sec:modeling:computationalaspects}. 

\subsection{Modeling}\label{sec:modeling:modeling}
Because counterfactual explanations (see Section~\ref{sec:foundations:counterfactuals}) proved to be an effective and useful explanation, we propose to use counterfactual explanations for explaining reject options of LVQ models\footnote{Although we focus on reject options for LVQ models, our proposed modeling is applicable to other models and reject options as well.} (see Section~\ref{sec:foundations:lvq}). Therefore, a counterfactual explanation of a reject provides the user with actionable feedback of what to change in order to be not rejected. Furthermore, such an explantion also communicates why the model is too uncertain for making a prediction in this particular case.

Since there exist evidence that people prefer low complexity (i.e. ``simple'') explanations, we are looking for sparse counterfactuals. Similar to~\refdef{def:counterfactual}, we phrase a counterfactual explanation $\xcf$ of a given input $\xorig$ as the following optimization problem:
\begin{equation}\label{eq:cfreject:opt}
\begin{split}
&\underset{\xcf\,\in\,\RN^\dimsym}{\min}\;\pnorm{\xorig - \xcf}_1 \\
\text{s.t. }& r(\xcf) \geq \threshold
\end{split}
\end{equation}
where $r(\cdot)$ denotes the specific reject option and the $\lOne$-norm objective is supposed to yield a sparse and hence a ``low complexity explanation''.

\subsection{Computational Aspects of LVQ Reject Options}\label{sec:modeling:computationalaspects}
In the following, we propose algorithms which solve convex optimizations problems, for efficiently computing counterfactual explanations of LVQ rejects -- whereby we consider each of the three reject options from Section~\ref{sec:foundations:rejectoptions} separately.
For the purpose of readability, we moved all proofs and derivations to Appendix~\ref{sec:appendix:counterfactuals}.

\subsubsection{Relative Similarity}
In case of the relative similarity reject option~\refeq{eq:relsim}, the optimization problem~\refeq{eq:cfreject:opt} can be solved by using a divide \& conquer approach, where we have to solve a bunch of convex quadratic programs of the following form:
\begin{equation}\label{eq:cfreject:relsim:opt}
\begin{split}
&\underset{\xcf\,\in\,\RN^\dimsym}{\min}\;\pnorm{\xorig - \xcf}_1 \\
\text{s.t. }& \xcf^\top\distmat\xcf + \q_j^\top\xcf + c_j \leq 0 \quad \forall\,\vec{\prototype}_j \in \set{P}(\protolabel_{+})
\end{split}
\end{equation}
where $\set{P}(\protolabel_{+})$ denotes a set of prototypes (see Appendix~\ref{sec:appendix:counterfactuals:relsim} for details).
Note that convex quadratic programs can be solved efficiently~\cite{Boyd2004}.

The complete algorithm for computing a counterfactual explanation is given in Algorithm~\ref{algo:cfreject:relsim}.
\begin{algorithm}
\caption{Counterfactual under the relative similarity reject option}\label{algo:cfreject:relsim}
\textbf{Input:} Original input $\xorig$, reject threshold $\threshold$, the LVQ model\\
\textbf{Output:} Counterfactual $\xcf$
\begin{algorithmic}[1]
 \State $\xcf = \vec{0}$ \Comment{Initialize dummy solution}
 \State $z = \infty$	\Comment{Sparsity of the best solution so far}
  \For{$\vec{\prototype}_+ \in \set{P}$} \Comment{Loop over every possible prototype}
 	\State Solving \refeq{eq:cfreject:relsim:opt} yields a counterfactual ${\xcf}{_{*}}$
 	\If{$\pnorm{\xorig - {\xcf}{_{*}}}_1 < z$} \Comment{Keep this counterfactual if it is sparser than the currently ``best'' counterfactual}
 		\State $z=\pnorm{\xorig - {\xcf}{_{*}}}_1$
 		\State $\xcf = {\xcf}{_{*}}$
 	\EndIf
 \EndFor
\end{algorithmic}
\end{algorithm}

\subsubsection{Distance to Decision Boundary}
Similar to the relative similarity reject option, we again use a divide \& conquer approach for solving~\refeq{eq:cfreject:opt}. But in contrast to the relative similarity reject option, we have to solve a bunch of linear programs only, which can be solved even faster than convex quadratic programs~\cite{Boyd2004}:
\begin{equation}\label{eq:cfreject:distdecbound:opt}
\begin{split}
&\underset{\xcf\,\in\,\RN^\dimsym}{\min}\;\pnorm{\xorig - \xcf}_1 \\
\text{s.t. }& \q_{j}^\top\xcf + c_{j} \geq 0 \quad \forall\,\vec{\prototype}_j\in \set{P}(\protolabel_{+})
\end{split}
\end{equation}
where $\set{P}(\protolabel_{+})$ denotes a set of prototypes (see Appendix~\ref{sec:appendix:counterfactuals:decdisbound} for details).

The final algorithm for computing a counterfactual explanation is equivalent to Algorithm~\ref{algo:cfreject:relsim} for the relative similarity reject option, except that instead of solving~\refeq{eq:cfreject:relsim:opt} in line 4, we have to solve~\refeq{eq:cfreject:distdecbound:opt}.

\subsubsection{Probabilistic Certainty Measure}
In case of the probabilistic certainty measure as a reject option~\refeq{eq:probareejct}, we have that:
\begin{equation}
\begin{split}
&\rprob(\x) \geq \threshold \\
&\Leftrightarrow \underset{\y\,\in\setY}\max\;p(\y\mid\x) \geq \threshold\\
&\Leftrightarrow \exists\,i\in\setY:\;p(\y=i\mid\x) \geq \threshold
\end{split}
\end{equation}
Applying the divide \& conquer paradigm over $i\in\setY$ for solving~\refeq{eq:cfreject:opt}, yields optimization problems of the following form:

\begin{equation}\label{eq:cfreject:proba:opt1}
\begin{split}
&\underset{\xcf\,\in\,\RN^\dimsym}{\min}\;\pnorm{\xorig - \xcf}_1 \\
\text{s.t. }& \sum_{j\neq i} \exp\Bigg(\alpha \frac{\prototypedist_{i,j}(\xcf,\protom) - \prototypedist_{i,j}(\xcf,\protop)}{\prototypedist_{i,j}(\xcf,\protom) + \prototypedist_{i,j}(\xcf,\protop)} + \beta\Bigg) + 1 - \frac{1}{\threshold} \leq 0
\end{split}
\end{equation}

While we could solve~\refeq{eq:cfreject:proba:opt1} directly -- yielding Algorithm~\ref{algo:cfreject:proba1} --, it is a rather difficult optimization problem because of its lack of any structure like convexity -- e.g. general (black-box) solvers might be the only applicable choice. We therefore, additionally, propose a convex approximation where we can still guarantee feasibility (i.e. validity of the final counterfactual) at the price of losing closeness -- i.e. we might not find the sparsest possible counterfactual, although finding a global optimum of~\refeq{eq:cfreject:proba:opt1} might be difficult as well.
\begin{algorithm}
\caption{Counterfactual under the probabilistic certainty reject option}\label{algo:cfreject:proba1}
\textbf{Input:} Original input $\xorig$, reject threshold $\threshold$, the LVQ model\\
\textbf{Output:} Counterfactual $\xcf$
\begin{algorithmic}[1]
 \State $\xcf = \vec{0}$ \Comment{Initialize dummy solution}
 \State $z = \infty$	\Comment{Sparsity of the best solution so far}
  \For{$i\in\setY$} \Comment{Loop over every possible class}
 	\State Solving \refeq{eq:cfreject:proba:opt1} yields a counterfactual ${\xcf}{_{*}}$
 	\If{$\pnorm{\xorig - {\xcf}{_{*}}}_1 < z$} \Comment{Keep this counterfactual if it is sparser than the currently ``best'' counterfactual}
 		\State $z=\pnorm{\xorig - {\xcf}{_{*}}}_1$
 		\State $\xcf = {\xcf}{_{*}}$
 	\EndIf
 \EndFor
\end{algorithmic}
\end{algorithm}
Approximating the constraint in~\refeq{eq:cfreject:proba:opt1} yields a convex quadratic constraint, which then results in a convex quadratic program as a final approximation of~\refeq{eq:cfreject:proba:opt1} -- for details see Appendix~\ref{sec:appendix:counterfactuals:proba}:
\begin{equation}\label{eq:cfreject:proba:optconvexapprox}
\begin{split}
&\underset{\xcf\,\in\,\RN^\dimsym}{\min}\;\pnorm{\xorig - \xcf}_1 \\
\text{s.t. }& \x^\top\distmat\x + \q_j^\top\x + c_j \leq 0 \quad \forall\,\vec{\prototype}_j\in \set{P}(\protolabel_{i})
\end{split}
\end{equation}
Using the approximation~\refeq{eq:cfreject:proba:optconvexapprox} requires us to iterate over every possible prototype, every possible class different from the $i$-th class, and finally over every possible class -- i.e. finally yielding Algorithm~\ref{algo:cfreject:proba2}.
\begin{algorithm}
\caption{Counterfactual under the probabilistic certainty reject option -- Approximation}\label{algo:cfreject:proba2}
\textbf{Input:} Original input $\xorig$, reject threshold $\threshold$, the LVQ model\\
\textbf{Output:} Counterfactual $\xcf$
\begin{algorithmic}[1]
 \State $\xcf = \vec{0}$ \Comment{Initialize dummy solution}
 \State $z = \infty$	\Comment{Sparsity of the best solution so far}
 \For{$i\in\setY$} \Comment{Loop over every possible class}
 	\For{$j\in\setY\setminus\{i\}$} \Comment{Loop over all other classes}
		\For{$\vec{\prototype}_i \in \set{P}$} \Comment{Loop over every possible prototype}	 	
 			\State Solving \refeq{eq:cfreject:proba:optconvexapprox} yields a counterfactual ${\xcf}{_{*}}$
 			\If{$\pnorm{\xorig - {\xcf}{_{*}}}_1 < z$} \Comment{Keep this counterfactual if it is sparser than the currently ``best'' counterfactual}
 				\State $z=\pnorm{\xorig - {\xcf}{_{*}}}_1$
 				\State $\xcf = {\xcf}{_{*}}$
 			\EndIf
 		\EndFor	
 	\EndFor
 \EndFor
\end{algorithmic}
\end{algorithm}
Although our proposed approximation~\refeq{eq:cfreject:proba:optconvexapprox} can be computed quite fast (because it is a convex quadratic program), it comes at the price of ``drastically'' increasing the complexity of the final divide \& conquer algorithm. While we have to solve $|\setY|$ optimization problems~\refeq{eq:cfreject:proba:opt1} in Algorithm~\ref{algo:cfreject:proba1}, we get a quadratic complexity (quadratic in the number of classes) for our proposed convex approximation (Algorithm~\ref{algo:cfreject:proba2}):
\begin{equation}
|\setY|^2\cdot P_N - |\setY|\cdot P_N
\end{equation}
where $P_N$ denotes the number of prototypes per class used in the pair-wise classifiers. This quadratic complexity could become a problem in case of a large number of classes and a large number of prototypes per class.

To summarize, we propose two divide \& conquer algorithms for computing counterfactual explanations of the probabilistic certainty reject option~\refeq{eq:probareejct}. In Algorithm~\ref{algo:cfreject:proba1}, we have to solve a rather complicated (i.e. unstructured) optimization problem, but we have to do this only a few times, whereas in Algorithm~\ref{algo:cfreject:proba2} we have to solve many convex quadratic programs. Although we have to solve many more optimization problems in Algorithm~\ref{algo:cfreject:proba2} than in Algorithm~\ref{algo:cfreject:proba2}, solving the optimization problems in Algorithm~\ref{algo:cfreject:proba2} is much easier and it is also possible to easily extend the optimization problems with additional constraints, such as plausibility constraints as proposed in~\cite{PlausibleCounterfactuals}. We think that in general, both algorithms have their areas of application and that practitioners should choose between them depending on their needs and the specific scenario.

\section{Experiments}\label{sec:experiments}
We empirically evaluate all our proposed algorithms for computing counterfactual explanations of different reject options (see Sections~\ref{sec:foundations:lvq},~\ref{sec:foundations:rejectoptions}) for GMLVQ on several different data sets. In our empirical evaluation, we consider two different properties:
\begin{itemize}
\item We evaluate algorithmic properties like sparsity, validity (in case of black-box solvers) and overlap (degree of agreement) between counterfactuals computed in different ways.
\item We also evaluate whether counterfactual explanations of reject (as proposed in this work) are able to find and explain known ground truth reasons for rejects -- i.e. evaluating the goodness of the explanations themselves.
\end{itemize}
All experiments are implemented in Python and the source code is available on GitHub\footnote{\url{https://github.com/andreArtelt/explaining_lvq_reject}}.

\subsection{Data Sets}
We consider the following data sets for our empirical evaluation -- all data sets are scaled and standardized except when using Algorithm~\ref{algo:cfreject:proba2}\footnote{In all cases, except for Algorithm~\ref{algo:cfreject:proba2}, standardizing the data is necessary to get the methods working (i.e. avoiding numerical problems) -- however, for Algorithm~\ref{algo:cfreject:proba2} the opposite is true: without standardizing the method works fine but if the data is standardized, the method does not work at all -- i.e. we observe numerical problems.}:

\subsubsection{Wine}
The ``Wine data set''~\cite{winedata} is used for predicting the cultivator of given wine samples based on their chemical properties. The data set contains $178$ samples and $13$ numerical features such as alcohol, hue and color intensity.

\subsubsection{Breast cancer}
The ``Breast Cancer Wisconsin (Diagnostic) Data Set''~\cite{breastcancer} is used for classifying breast cancer samples into benign and malignant. The data set contains $569$ samples and $30$ numerical features such as area, smoothness and compactness.

\subsubsection{Flip}
This data set is used for the prediction of fibrosis. The set consists of samples of $118$ patients and $12$ numerical features such as blood glucose, BMI and total cholesterol, and was provided by the Department of Gastroenterology, Hepatology and Infectiology of the University Magdeburg~\cite{flipdata}.
As the data set contains some rows with missing values, we chose to replace these missing values with the corresponding feature mean.

\subsubsection{T21}
This data set is used for early diagnosis of chromosomal abnormalities, such as trisomy 21, in pregnant women. The data set consists of $18$ numerical features such as heart rate and weight, and contains over $50000$ samples but only $0.8$ percent abnormal samples (e.g. cases of trisomy 21) -- i.e. it is highly imbalanced. It was collected by the Fetal Medicine Centre at King's College Hospital and University College London Hospital in London~\cite{t21dataset}.

\subsection{Setup}
For finding the best model parameters we perform a grid search on the GMLVQ hyperparameters and reject option model's rejection thresholds. In order to reduce the risk of overfitting, we cross validate the GMLVQ models' accuracy and rejection rates on the different reject thresholds using a 5-fold cross validation.

For each GMLVQ model hyperparameterization, the rejection rates for each of the thresholds is computed as well as the impact of this on the accuracy of the model. Based on these rejection rates and accuracies, the accuracy-rejection curve (ARC)~\cite{nadeem2009accuracy} can be computed. The area under the ARC measure (AU-ARC) gives an indication of how well the reject model performs given the GMLVQ model type and its hyperparameterization.
%By generating these scores over .
For each combination of data set, GMLVQ model type, and reject option method, we can then determine the best GMLVQ model parameters and rejection threshold. We do this by selecting the GMLVQ hyperparameters with the highest accuracy. Another alternative method would the inclusion of the AU-ARC as an indication of how well the model performs. Furthermore, we could combine these two measures by using some type of weighting strategy.

Following the selection of the best GMLVQ hyperparameters, the best rejection threshold is found by determining the ``optimal'' threshold in the ARC by finding the so-called knee-point using the Kneedle algorithm~\cite{satopaa2011kneedle}. In reality, the truly optimal threshold depends on the data set and the real-world application in which it is used. However, for the purposes of our evaluation, we are primarily interested in those GMLVQ hyperparameters and reject option thresholds that allow us to compare the overall performance and impact of the different methods for computing counterfactuals.
Once the optimal hyperparameters have been obtained, the counterfactuals of the different reject options using different algorithms, can be computed, compared, and evaluated. 
By finding the optimal hyperparameters for each model, our evaluation of the different model types will be less dependent on potentially poor hyperparamerization.

We evaluate two properties separately: Algorithmic properties like sparsity and goodness of the explanations.
We run the experiments (5-fold cross validation) for each data set, each reject option and each method for computing the counterfactuals -- i.e black-box solver for solving~\refeq{eq:cfreject:opt}\footnote{We use the penalty method (with equal weighting) to turn~\refeq{eq:cfreject:opt} into an unconstrained problem, solve it by using the Nelder-Mead black-box method.}, our proposed algorithms from Section~\ref{sec:modeling}) and the closest sample from the training data set which is not rejected (this ``naive'' way of computing a counterfactual serves as a baseline).

\subsubsection{Algorithmic Properties}
We evaluate the sparsity of the counterfactual explanations by using the $\lZero$-norm\footnote{For numerical reasons, we use a threshold at which we consider a floating point number to be equal to zero -- see provided source code for details.}. Since our methods and algorithms are guaranteed to output feasible (i.e. valid) counterfactuals, we only evaluate validity of the counterfactuals computed by the black-box solver. Furthermore, we also compare the pairwise overlap/agreement of the counterfactuals computed by the different methods -- i.e. how large is the overlap in the selected (non-zero) features of the counterfactuals computed by two different methods.

\subsubsection{Goodness of Counterfactual Explanations}
For evaluating the goodness of the counterfactuals, we create scenarios with ground truth as follows: For each data set, we select a random subset of features ($30\%$) and perturb these in the test set by adding Gaussian noise -- we then check which of these samples are rejected due to the noise (i.e. applying the reject option before and after applying the perturbation), and compute counterfactual explanations of these samples only.
We then evaluate for each counterfactual how many of the relevant features (from the known ground truth) are detected and included in the counterfactual explanation.

\subsection{Results \& Discussion}
When reporting the results, we use the following abbreviations: \textit{BbCfFeasibility} -- Feasibility of the counterfactuals computed by the black-box solver, in case of the probabilistic reject option~\refeq{eq:probareejct}, we report the results of using Algorithm~\ref{algo:cfreject:proba1} (the results for the ``true'' black-box solver can be found in Appendix~\ref{sec:appendix:experimens:goodnessexplanation}); \textit{BbCf} -- Counterfactuals computed by the black-box solver, \textit{TrainCf} -- Counterfactuals by selecting the closest sample from the training set which is not rejected; \textit{ClosestCf} -- Counterfactuals computed by our proposed algorithms.

Note that we round all values to two decimal points.

\subsubsection{Algorithmic Properties}
In Table~\ref{table:experimentresults:sparsity}, we report the mean sparsity (along with the variance) of the counterfactuals for different reject options and different data sets.
The results (mean and variance) for evaluating the number of overlapping features are reported in Table~\ref{table:experimentresults:overlap}.
\begin{table}
\caption{Algorithmic properties -- Mean sparsity (incl. variance) of different counterfactuals, smaller values are better.}
\centering
\footnotesize
\begin{tabular}{|c|c||c||c|c|c||}
 \hline
 & \textit{DataSet} & BbCfFeasibility & BbCf & TrainCf & ClosestCf \\
 \hline
 \multirow{5}{*}{\rotatebox[origin=l]{0}{$\text{Relative}\atop\text{Similarity}$~\refeq{eq:relsim}}}
 & Wine & $1.0 \pm 0.0$ & $11.14 \pm 1.41$ & $13.0 \pm 0.0$ & $5.68 \pm 11.36$ \\
 & Breast Cancer & $0.99 \pm 0.0$ & $27.64 \pm 4.38$ & $30.0 \pm 0.0$ & $12.69 \pm 52.31$ \\
 & t21 & $0.99 \pm 0.0$ & $16.14 \pm 1.91$ & $11.0 \pm 1.0$ & $4.39 \pm 12.79$ \\
 & Flip & $1.0 \pm 0.0$ & $10.52 \pm 1.38$ & $12.0 \pm 0.0$ & $4.07 \pm 8.02$ \\
 \hline\hline
 \multirow{5}{*}{\rotatebox[origin=l]{0}{$\text{Distance}\atop\text{DecisionBoundary}$~\refeq{eq:rejectdist}}}
 & Wine & $1.0 \pm 0.0$ & $10.98 \pm 1.12$ & $13.0 \pm 0.0$ & $2.8 \pm 9.13$ \\
 & Breast Cancer & $1.0 \pm 0.0$ & $27.5 \pm 2.78$ & $30.0 \pm 0.0$ & $4.16 \pm 48.19$ \\
 & t21 & $1.0 \pm 0.0$ & $16.02 \pm 1.43$ & $11.0 \pm 1.0$ & $2.12 \pm 5.19$ \\
 & Flip & $1.0 \pm 0.0$ & $10.31 \pm 1.2$ & $12.0 \pm 0.0$ & $1.88 \pm 3.11$ \\
 \hline\hline
 \multirow{5}{*}{\rotatebox[origin=l]{0}{$\text{Probabilistic}\atop\text{Certainty}$~\refeq{eq:probareejct}}}
 & Wine & $0.45 \pm 0.21$ & $13.0 \pm 0.0$ & $13.0 \pm 0.0$ & $12.67 \pm 0.22$ \\
 & Breast Cancer & $0.94 \pm 0.02$ & $30.0 \pm 0.0$ & $30.0 \pm 0.0$ & $26.04 \pm 60.5$ \\
 & t21 & $0.4 \pm 0.24$ & $17.39 \pm 0.64$ & $11.0 \pm 0.0 $ & $12.55 \pm 7.41$ \\
 & Flip & $0.4 \pm 0.24$ & $11.75 \pm 0.23$ & $12.0 \pm 0.0$ & $11.5 \pm 1.47$ \\
 \hline
\end{tabular}
\label{table:experimentresults:sparsity}
\end{table}
\begin{table}
\caption{Algorithmic properties -- Mean overlap (incl. variance) between different counterfactuals.}
\centering
\footnotesize
\begin{tabular}{|c|c||c||c|c|c||}
 \hline
 & \textit{DataSet} & BbCfFeasibility & BbCf & TrainCf & ClosestCf\\
 \hline
 \multirow{5}{*}{\rotatebox[origin=l]{0}{$\text{Relative}\atop\text{Similarity}$~\refeq{eq:relsim}}}
 & Wine & $1.0 \pm 0.0$ & $4.68 \pm 7.26$ & $11.16 \pm 1.45$ & $5.27 \pm 9.02$ \\
 & Breast Cancer & $0.99 \pm 0.0$ & $10.86 \pm 54.34$ & $27.34 \pm 5.18$ & $11.37 \pm 59.55$ \\
 & t21 & $0.99 \pm 0.0$ & $3.92 \pm 10.5$ & $9.61 \pm 2.14$ & $3.52 \pm 9.42$ \\
 & Flip & $1.0 \pm 0.0$ &  $3.8 \pm 6.94$ & $10.11 \pm 1.55$ & $4.19 \pm 8.73$  \\
 \hline\hline
 \multirow{5}{*}{\rotatebox[origin=l]{0}{$\text{Distance}\atop\text{DecisionBoundary}$~\refeq{eq:rejectdist}}}
 & Wine & $1.0 \pm 0.0$ & $2.18 \pm 5.14$ & $11.08 \pm 1.13$ & $2.39 \pm 7.21$ \\
 & Breast Cancer & $1.0 \pm 0.0$ & $4.66 \pm 51.26$ & $27.83 \pm 3.13$ & $4.85 \pm 56.74$ \\
 & t21 & $1.0 \pm 0.0$ & $1.67 \pm 1.75$ & $10.0 \pm 1.87$ & $1.53 \pm 1.39$\\
 & Flip & $1.0 \pm 0.0$ & $1.59 \pm 1.17$ & $9.95 \pm 1.57$ & $1.68 \pm 1.35$ \\
 \hline\hline
 \multirow{5}{*}{\rotatebox[origin=l]{0}{$\text{Probabilistic}\atop\text{Certainty}$~\refeq{eq:probareejct}}}
 & Wine & $0.45 \pm 0.21$ & $11.09 \pm 2.41$ & $11.0 \pm 2.1$ & $12.83 \pm 0.14$ \\
 & Breast Cancer & $0.94 \pm 0.02$ & $24.75 \pm 8.77$ & $22.0 \pm 1.43$ & $29.42 \pm 0.93$ \\
 & t21 & $0.4 \pm 0.24$ & $9.26 \pm 1.8$ & $8.47 \pm 2.23$ & $11.54 \pm 9.55$  \\
 & Flip & $0.4 \pm 0.24$ & $11.71 \pm 0.29$ & $11.69 \pm 0.34$ & $12.0 \pm 0.0$ \\
 \hline
\end{tabular}
\label{table:experimentresults:overlap}
\end{table}
We observe that our proposed methods for computing counterfactual explanation of reject options is consistently able to compute very sparse (i.e. low complexity) counterfactuals -- however, the variance is often very large which suggests that there exist a few outliers in the data set for which it is not possible to compute a sparse counterfactuals.
As it was to be expected, we observe the worst performance when choosing a sample from the training set as a counterfactual. The counterfactuals computed by a black-box solver are often a bit better than those from the training set but still far away from the counterfactuals computed by our proposed algorithms.
While the black-box solver works quite well in case of the relative similarity and distance to decision boundary reject options, the performance drops significantly (for many but not all data sets) in case of the probabilistic certainty reject option. We think this might be due to the increased complexity of the reject option, compared to the other two reject options which have much simpler mathematical form. For this reject option, our proposed algorithm is still able to consistently yield the sparsest counterfactuals but the difference to other counterfactuals is not that significant like it is the case for the other two reject options. 

Concerning the feature overlap of the different counterfactuals, we observe that usually all methods agree upon the same counterfactual features.

\subsubsection{Goodness of Counterfactual Explanations}
The mean recall (along with the variance) of recovered (ground truth) relevant features for different counterfactuals, data sets and reject options, is given in Table~\ref{table:experimentresults:goodnessexplanations}. 
\begin{table}
\caption{Goodness of counterfactual explanations -- Mean and variance recall of identified relevant features (larger numbers are better).}
\centering
\footnotesize
\begin{tabular}{|c|c||c||c|c|c||}
 \hline
 & \textit{DataSet} & BbCfFeasible & BbCf & TrainCf & Cf\\
 \hline
 \multirow{5}{*}{\rotatebox[origin=l]{0}{$\text{Relative}\atop\text{Similarity}$~\refeq{eq:relsim}}}
 & Wine & $1.0 \pm 0.0$ & $1.0 \pm 0.0$ & $1.0 \pm 0.0$ & $1.0 \pm 0.0$ \\
 & Breast Cancer & $1.0 \pm 0.0$ & $1.0 \pm 0.0$ & $1.0 \pm 0.0$ & $0.99 \pm 0.01$ \\
 & t21 & $1.0 \pm 0.0$ & $1.0 \pm 0.0$ & $1.0 \pm 0.0$ & $0.98 \pm 0.0$ \\
 & Flip & $1.0 \pm 0.0$ & $1.0 \pm 0.0$ & $1.0 \pm 0.0$ & $0.96 \pm 0.02$ \\
 \hline\hline
 \multirow{5}{*}{\rotatebox[origin=l]{0}{$\text{Distance}\atop\text{DecisionBoundary}$~\refeq{eq:rejectdist}}}
 & Wine & $0.8 \pm 0.16$ & $1.0 \pm 0.0$ & $1.0 \pm 0.0$ & $0.72 \pm 0.15$ \\
 & Breast Cancer & $1.0 \pm 0.0$ & $1.0 \pm 0.0$ & $1.0 \pm 0.0$ & $0.45 \pm 0.18$ \\
 & t21 & $0.6 \pm 0.24$ & $1.0 \pm 0.0$ & $1.0 \pm 0.0$ & $0.63 \pm 0.15$ \\
 & Flip & $1.0 \pm 0.0$ & $1.0 \pm 0.0$ & $1.0 \pm 0.0$ & $0.62 \pm 0.22$ \\
 \hline\hline
 \multirow{5}{*}{\rotatebox[origin=l]{0}{$\text{Probabilistic}\atop\text{Certainty}$~\refeq{eq:probareejct}}}
 & Wine & $0.8 \pm 0.16$ & $1.0 \pm 0.0$ & $1.0 \pm 0.0$ & $1.0 \pm 0.0$  \\
 & Breast Cancer & $0.8 \pm 0.16$ & $1.0 \pm 0.0$ & $1.0 \pm 0.0$ & $1.0 \pm 0.0$  \\
 & t21 & $0.85 \pm 0.02$ & $1.0 \pm 0.0$ & $1.0 \pm 0.0$ & $0.94 \pm 0.01$ \\
 & Flip & $0.73 \pm 0.06$ & $1.0 \pm 0.0$ &  $1.0 \pm 0.0$ & $0.75 \pm 0.19$\\
 \hline
\end{tabular}
\label{table:experimentresults:goodnessexplanations}
\end{table}
We observe that all methods are able to identify the relevant features that caused the reject. However, the counterfactuals computed by our proposed algorithms sometimes miss a few relevant features which is most likely due to the fact that these methods are designed to find the sparsest possible counterfactual -- this indicates that sparsity alone is not always sufficient for coming up with the ``best'' explanation. % Results vary quite a lot for different (non-optimal) thresholds :/

\section{Summary \& Conclusion}\label{sec:conclusion}
In this work we proposed to explain reject options of LVQ models by means of counterfactual explanations. We considered three popular reject options and for each, proposed (extendable) modelings and algorithms for efficiently computing counterfactual explanations under the particular reject option. We empirically evaluated all our proposed methods under different aspects -- in particular, we demonstrated that our algorithms delivers sparse (i.e. ``low-complexity'') explanations, and that counterfactual explanations in general seem to be able to detect and highlight relevant features in scenarios where the ground truth is known.

Although our proposed idea of using counterfactual explanations for explaining rejects is rather general, our proposed methods and algorithms are completely tailored towards LVQ models and thus not applicable to other ML models. Therefore, it would be of interest to see other, either model specific or more general methods for computing counterfactual explanations of reject under other ML models.

Our evaluation focused on algorithmic properties such as sparsity and feature relevances for assessing the goodness of the computed counterfactuals. However, it is still unclear how and if these kinds of explanations of reject are useful and helpful to humans -- since it is difficult to implement ``human usefulness'' as a scoring function, a proper user study for evaluating the usefulness is necessary.

We leave these aspects as future work.


%\bibliographystyle{ACM-Reference-Format}
\bibliographystyle{splncs04}
\bibliography{bibliography.bib}

\appendix
\section{Proofs \& Derivations}\label{sec:appendix}
\subsection{Counterfactual Explanations}\label{sec:appendix:counterfactuals}
\subsubsection{Relative Similarity}\label{sec:appendix:counterfactuals:relsim}
In order for a sample $\x\in\RN^\dimsym$ to be classified, it must hold that:
\begin{equation}\label{appendix:eq:relsim}
\relsim(\x) \geq \threshold
\end{equation}
Using the shorter notation $\prototypedist^{+}(\x)=\prototypedist(\x,\protop)$ and  $\prototypedist^{-}(\x)=\prototypedist(\x,\protom)$, respectively, where $\protop$ refers to the closest prototype and $\protom$ revers to the closest one belongs to a different class, this further translates into:
\begin{equation}\label{eq:relsim:constraint}
\begin{split}
&\relsim(\x) \geq \threshold \\
&\Leftrightarrow\frac{\prototypedist^{-}(\x) - \prototypedist^{+}(\x)}{\prototypedist^{-}(\x) + \prototypedist^{+}(\x)} \geq \threshold\\
&\Leftrightarrow{\prototypedist}^{-}(\x) - {\prototypedist}^{+}(\x) \geq \left({\prototypedist}^{-}(\x) + {\prototypedist}^{+}(\x)\right)\threshold\\
&\Leftrightarrow{\prototypedist}^{-}(\x) - {\prototypedist}^{-}(\x)\threshold \geq {\prototypedist}^{+}(\x) + {\prototypedist}^{+}(\x)\threshold \\
&\Leftrightarrow (1-\threshold){\prototypedist}^{-}(\x) - (1+\threshold){\prototypedist}^{+}(\x) \geq 0
\end{split}
\end{equation}
Assuming that the closest prototype $\vec{\prototype}_{+}$ is fixed, we can rewrite~\refeq{eq:relsim:constraint} as follows:
\begin{equation}\label{eq:relsim:constraint:new}
(1-\threshold){\prototypedist}(\x, \vec{\prototype}_j) - (1+\threshold){\prototypedist}(\x, \vec{\prototype}_{+}) \geq 0 \quad \forall\,\vec{\prototype}_j \in \set{P}(\protolabel_{+})
\end{equation}
where $\set{P}(\protolabel_{+})$ denotes the set of all prototypes that are not labeled as $\protolabel_{+}$.
The intuition behind the translation of~\refeq{eq:relsim:constraint} into~\refeq{eq:relsim:constraint:new} is that, if~\refeq{eq:relsim:constraint} holds for all possible $\vec{\prototype}_{-}$, then also for the particular choice of $\vec{\prototype}_{-}$ in~\refeq{appendix:eq:relsim}.

These constraints can be rewritten as the following convex quadratic constraints -- for a given $\vec{\prototype}_j$:
\begin{equation}
\begin{split}
&(1-\threshold){\prototypedist}(\x, \vec{\prototype}_j) - (1+\threshold){\prototypedist}(\x, \vec{\prototype}_{+}) \geq 0 \\
&\Leftrightarrow {\prototypedist}(\x, \vec{\prototype}_j) - {\prototypedist}(\x, \vec{\prototype}_{+}) - \threshold\Big({\prototypedist}(\x, \vec{\prototype}_j) + {\prototypedist}(\x, \vec{\prototype}_{+})\Big) \geq 0 \\
&\Leftrightarrow 2(\vec{\prototype}_{+}^\top\distmat - \vec{\prototype}_j^\top\distmat)\x + \vec{\prototype}_j^\top\distmat\vec{\prototype}_j - \vec{\prototype}_{+}^\top\distmat\vec{\prototype}_{+} -\\&\quad \threshold\Big(2\x^\top\distmat\x - 2(\vec{\prototype}_j^\top\distmat + \vec{\prototype}_{+}^\top\distmat)\x + \vec{\prototype}_j^\top\distmat\vec{\prototype}_j + \vec{\prototype}_{+}^\top\distmat\vec{\prototype}_{+}\Big) \geq 0 \\
&\Leftrightarrow  -2(\vec{\prototype}_{+}^\top\distmat - \vec{\prototype}_j^\top\distmat)\x - \vec{\prototype}_j^\top\distmat\vec{\prototype}_j + \vec{\prototype}_{+}^\top\distmat\vec{\prototype}_{+} +\\&\quad \threshold\Big(2\x^\top\distmat\x - 2(\vec{\prototype}_j^\top\distmat + \vec{\prototype}_{+}^\top\distmat)\x + \vec{\prototype}_j^\top\distmat\vec{\prototype}_j + \vec{\prototype}_{+}^\top\distmat\vec{\prototype}_{+}\Big) \leq 0 \\
&\Leftrightarrow \x^\top\distmat\x + \q_j^\top\x + c_j \leq 0
\end{split}
\end{equation}
where
\begin{equation}
\q_j^\top = \left(-\frac{1}{\threshold} - 1\right)\vec{\prototype}_{+}^\top\distmat + \left(\frac{1}{\threshold} - 1\right)\vec{\prototype}_{j}^\top\distmat
\end{equation}
\begin{equation}
c_j = \frac{1}{2}\Bigg(\left(1 - \frac{1}{\threshold}\right)\vec{\prototype}_j^\top\distmat\vec{\prototype}_j + \left(1 + \frac{1}{\threshold}\right)\vec{\prototype}_{+}^\top\distmat\vec{\prototype}_{+}\Bigg)
\end{equation}
We therefore get the following convex quadratic optimization problem -- note that convex quadratic programs can be solved efficiently~\cite{Boyd2004}:
\begin{equation}\label{eq:relsim:opt}
\begin{split}
&\underset{\xcf\,\in\,\RN^\dimsym}{\min}\;\pnorm{\xorig - \xcf}_1 \\
\text{s.t. }& \xcf^\top\distmat\xcf + \q_j^\top\xcf + c_j \leq 0 \quad \forall\,\vec{\prototype}_j \in \set{P}(\protolabel_{+})
\end{split}
\end{equation}
We simply solve this optimization problem~\refeq{eq:relsim:opt} for all possible target prototypes and select the counterfactual which is the closest to the original sample -- since every possible prototype could be a potential closest prototype $\vec{\prototype}_{+}$, we have to solve as many optimization problems as we have prototypes. However, one could introduce some kind of early stopping by adding an additional constraint on the distance of the counterfactual to the original sample -- i.e. use the currently best known solution as a upper bound on the objective, which might result in an infeasible program and hence will be aborted quickly. Alternatively, one could solve the different optimization problems in parallel because they are no dependencies between them.

\subsubsection{Distance to decision boundary}\label{sec:appendix:counterfactuals:decdisbound}
In order for a sample $\x\in\RN^\dimsym$ to be accepted (i.e. not being rejected), it must hold that:
\begin{equation}\label{eq:distdecbound}
\rdist(\x) \geq \threshold
\end{equation}
This further translates into:
\begin{equation}\label{eq:distdecbound:constraint}
\begin{split}
&\rdist(\x) \geq \threshold\\
&\Leftrightarrow\frac{|\prototypedist^{+}(\x) - \prototypedist^{-}(\x)|}{2\pnorm{\vec{\prototype}_{+} - \vec{\prototype}_{-}}_2^2} \geq \threshold\\
&\Leftrightarrow |{\prototypedist}^{+}(\x) - {\prototypedist}^{-}(\x)| \geq 2\threshold\pnorm{\vec{\prototype}_{+} - \vec{\prototype}_{-}}_2^2\\
&\Leftrightarrow {\prototypedist}^{-}(\x) - {\prototypedist}^{+}(\x) - 2\threshold\pnorm{\vec{\prototype}_{+} - \vec{\prototype}_{-}}_2^2 \geq 0
\end{split}
\end{equation}
Assuming that the closest prototype $\vec{\prototype}_{+}$ is fixed, we get:
\begin{equation}\label{eq:distdecbound:constraint:new}
\prototypedist(\x, \vec{\prototype}_j) - {\prototypedist}(\x, \vec{\prototype}_{+}) - 2\threshold\pnorm{\vec{\prototype}_{+} - \vec{\prototype}_j}_2^2 \geq 0 \quad \forall\,\vec{\prototype}_j\in \set{P}(\protolabel_{+})
\end{equation}
where $\set{P}(\protolabel_{+})$ denotes the set of prototypes that are not labeled as $\protolabel_{+}$.
Again, the intuition behind this translation is that, if~\refeq{eq:distdecbound:constraint} holds for all possible $\vec{\prototype}_{-}$, then also for the particular choice of $\vec{\prototype}_{-}$ in~\refeq{eq:distdecbound}.

These constraints can be rewritten as the following linear constraints -- for a given $\vec{\prototype}_j$:
\begin{equation}
\begin{split}
&\prototypedist(\x, \vec{\prototype}_j) - {\prototypedist}(\x, \vec{\prototype}_{+}) - 2\threshold\pnorm{\vec{\prototype}_{+} - \vec{\prototype}_j}_2^2 \geq 0 \\
&\Leftrightarrow (\x - \vec{\prototype}_j)^\top\distmat(\x - \vec{\prototype}_j) - (\x - \vec{\prototype}_{+})^\top\distmat(\x - \vec{\prototype}_{+}) -\\&\quad 2\threshold(\vec{\prototype}_{+} - \vec{\prototype}_j)^\top\distmat(\vec{\prototype}_{+} - \vec{\prototype}_j) \geq 0\\
&\Leftrightarrow \x^\top\distmat\x - \x^\top\distmat\vec{\prototype}_j - \vec{\prototype}_j^\top\distmat\x + \vec{\prototype}_j^\top\distmat\vec{\prototype}_j - \x^\top\distmat\x + \x^\top\distmat\vec{\prototype}_{+} +\\&\quad \vec{\prototype}_{+}^\top\distmat\x - \vec{\prototype}_{+}^\top\distmat\vec{\prototype}_{+}  - 2\threshold(\vec{\prototype}_{+} - \vec{\prototype}_j)^\top\distmat(\vec{\prototype}_{+} - \vec{\prototype}_j) \geq 0 \\
&\Leftrightarrow \left(2\vec{\prototype}_{+}^\top\distmat - 2\vec{\prototype}_j^\top\distmat\right)\x + \vec{\prototype}_j^\top\distmat\vec{\prototype}_j - \vec{\prototype}_{+}^\top\distmat\vec{\prototype}_{+} -\\&\quad 2\threshold(\vec{\prototype}_{+} - \vec{\prototype}_j)^\top\distmat(\vec{\prototype}_{+} - \vec{\prototype}_j) \geq 0\\
&\Leftrightarrow \q_j^\top\x + c_j \geq 0
\end{split}
\end{equation}
where
\begin{equation}
\q_j^\top = 2\vec{\prototype}_{+}^\top\distmat - 2\vec{\prototype}_j^\top\distmat \quad\quad c_j = \vec{\prototype}_j^\top\distmat\vec{\prototype}_j - \vec{\prototype}_{+}^\top\distmat\vec{\prototype}_{+} - 2\threshold(\vec{\prototype}_{+} - \vec{\prototype}_j)^\top\distmat(\vec{\prototype}_{+} - \vec{\prototype}_j)
\end{equation}
Finally, we get the following linear optimization problem -- note that linear programs can be solved even faster than convex quadratic programs~\cite{Boyd2004}:
\begin{equation}
\begin{split}
&\underset{\xcf\,\in\,\RN^\dimsym}{\min}\;\pnorm{\xorig - \xcf}_1 \\
\text{s.t. }& \q_{j}^\top\xcf + c_{j} \geq 0 \quad \forall\,\vec{\prototype}_j\in \set{P}(\protolabel_{+})
\end{split}
\end{equation}
Again, we try all possible target prototypes $\vec{\prototype}_{+}$ and select the best counterfactual -- everything from the relative similarity case applies (see Section~\ref{sec:appendix:counterfactuals:relsim}).

\subsubsection{Probabilistic certainty measure}\label{sec:appendix:counterfactuals:proba}
In order for a sample $\x\in\RN^\dimsym$ to be classified (i.e. not being rejected), it must hold that:
\begin{equation}
\rprob(\x) \geq \threshold
\end{equation}
This further translates into:
\begin{equation}
\begin{split}
&\rprob(\x) \geq \threshold \\
&\Leftrightarrow \underset{\y\,\in\,\setY}\max\;p(\y\mid\x) \geq \threshold\\
&\Leftrightarrow \exists\,i\in\setY:\;p(\y=i\mid\x) \geq \threshold
\end{split}
\end{equation}
For the moment, we assume that $i$ is fixed -- i.e. using a divide \& conquer approach. It follows that:%Divide-Conquer approach!
\begin{equation}\label{eq:probcertainmeasure:constraint:begin}
\begin{split}
&p(\y=i\mid\x) \geq \threshold\\
&\Leftrightarrow \frac{1}{\sum_{j\neq i}\frac{1}{p(\y=i\mid (i, j), \x)} - (|\setY| - 2)} \geq \threshold\\
&\Leftrightarrow 1 \geq\threshold\left(\sum_{j\neq i}\frac{1}{p(\y=i\mid (i, j), \x)} - (|\setY| - 2)\right) \\
&\Leftrightarrow \sum_{j\neq i}\frac{1}{p(\y=i\mid (i, j), \x)} - c \leq 0
\end{split}
\end{equation}
where
\begin{equation}
c = |\setY| - 2 + \frac{1}{\threshold} % > 0
\end{equation}
Further simplifications of~\refeq{eq:probcertainmeasure:constraint:begin} yield:
\begin{equation}\label{eq:probcertainmeasure:constraint}
\begin{split}
&\sum_{j\neq i}\frac{1}{p(\y=i\mid (i, j), \x)} - c \leq 0 \\
&\Leftrightarrow \sum_{j\neq i}1 + \exp(\alpha \cdot \rel_{i,j}(\x) + \beta) - c \leq 0\\
&\Leftrightarrow \sum_{j\neq i}\exp(\alpha \cdot \rel_{i,j}(\x) + \beta) + c' \leq 0 \\
&\Leftrightarrow \sum_{j\neq i} \exp\Bigg(\alpha \frac{\prototypedist_{i,j}^{-}(\xcf) - \prototypedist_{i,j}^{+}(\xcf)}{\prototypedist_{i,j}^{-}(\xcf) + \prototypedist_{i,j}^{+}(\xcf)} + \beta\Bigg) + c' \leq 0
\end{split}
\end{equation}
where
\begin{equation}
\begin{split}
c' &= |\setY|-1-c\\
&= |\setY| - 1 - |\setY| + 2 - \frac{1}{\threshold}\\
&= 1 - \frac{1}{\threshold}
\end{split}
\end{equation}
We therefore get the following optimization problem:
\begin{equation}\label{eq:probcertainmeasure:opt}
\begin{split}
&\underset{\xcf\,\in\,\RN^\dimsym}{\min}\;\pnorm{\xorig - \xcf}_1 \\
\text{s.t. }& \sum_{j\neq i} \exp\Bigg(\alpha \frac{\prototypedist_{i,j}^{-}(\x) - \prototypedist_{i,j}^{+}(\x)}{\prototypedist_{i,j}^{-}(\x) + \prototypedist_{i,j}^{+}(\x)} + \beta\Bigg) + c' \leq 0
\end{split}
\end{equation}
Because of the divide \& conquer paradigm, we would have to try all possible target classes $i\in\setY$ and finally select the one yielding the lowest objective -- i.e. the closest counterfactual. While one could do this using constraint~\refeq{eq:probcertainmeasure:constraint}, the optimization would be rather complicated because the constraint is not convex and rather ``ugly'' -- although it could be tackled by an evolutionary optimization method. However, the number of optimization problems we have to solve is rather small -- it is equal to the number of classes.

We therefore, additionally, propose a surrogate constraint which captures the same ``meaning/intuition'' as~\refeq{eq:probcertainmeasure:constraint} does, but is easier to optimize over -- however, note that by using a surrogate instead of the original constraint~\refeq{eq:probcertainmeasure:constraint} we give up closeness which, in our opinion, would be acceptable if the solutions stay somewhat close to each other\footnote{Furthermore, in case of additional plausibility \& actionability constraints, closeness becomes even less important.}. We try out and compare both approaches in the experiments (see Section~\ref{sec:experiments}).

First, we apply the natural logarithm to~\refeq{eq:probcertainmeasure:constraint} and then bound it by using the maximum:
\begin{equation}\label{eq:probcertainmeasure:constraint:surrogate}
\begin{split}
&\log\left(\sum_{j\neq i} \exp\Bigg(\alpha \frac{\prototypedist_{i,j}^{-}(\x) - \prototypedist_{i,j}^{+}(\x)}{\prototypedist_{i,j}^{-}(\x) + \prototypedist_{i,j}^{+}(\x)} + \beta\Bigg)\right) \\&\leq  \max_{j\neq i}\left(\alpha \frac{\prototypedist_{i,j}^{-}(\x) - \prototypedist_{i,j}^{+}(\x)}{\prototypedist_{i,j}^{-}(\x) + \prototypedist_{i,j}^{+}(\x)} + \beta\right) + \log\left(|\setY|-1\right)
\end{split}
\end{equation}
We therefore approximate the constraint~\refeq{eq:probcertainmeasure:constraint} by using~\refeq{eq:probcertainmeasure:constraint:surrogate}, which yields the following constraint:
\begin{equation}
\max_{j\neq i}\left(\alpha \frac{\prototypedist_{i,j}^{-}(\x) - \prototypedist_{i,j}^{+}(\x)}{\prototypedist_{i,j}^{-}(\x) + \prototypedist_{i,j}^{+}(\x)} + \beta\right) \leq \log(-c') - \log(|\setY|-1)
\end{equation}
Note that, in theory it could happen that $-c'\leq 0$ -- we fix this by simply taking $\max(-c', \epsilon)$, which results in a feasible solution but the approximation gets a bit worse.

Assuming that the maximum $j$ is fixed, we get the following constraint:
\begin{equation}
\frac{\prototypedist_{i,j}^{-}(\x) - \prototypedist_{i,j}^{+}(\x)}{\prototypedist_{i,j}^{-}(\x) + \prototypedist_{i,j}^{+}(\x)}  \leq \gamma
\end{equation}
where
\begin{equation}
\gamma = \frac{\log(-c')  - \log(|\setY|-1) - \beta}{\alpha}
\end{equation}
Further simplifications reveal that:
\begin{equation}\label{eq:probcertainmeasure:constraint:convexapprox}
\begin{split}
&\frac{\prototypedist_{i,j}^{-}(\x) - \prototypedist_{i,j}^{+}(\x)}{\prototypedist_{i,j}^{-}(\x) + \prototypedist_{i,j}^{+}(\x)}  \leq \gamma \\
&\Leftrightarrow {\prototypedist}_{i,j}^{-}(\x) - {\prototypedist}_{i,j}^{+}(\x) \leq \Big({\prototypedist}_{i,j}^{-}(\x) + {\prototypedist}_{i,j}^{+}(\x)\Big)\gamma \\
&\Leftrightarrow {\prototypedist}_{i,j}^{-}(\x) - {\prototypedist}_{i,j}^{+}(\x) - \gamma{\prototypedist}_{i,j}^{-}(\x) - \gamma{\prototypedist}_{i,j}^{+}(\x) \leq 0 \\
&\Leftrightarrow (1-\gamma){\prototypedist}_{i,j}^{-}(\x) - (1+\gamma){\prototypedist}_{i,j}^{+}(\x) \leq 0 \\
&\Leftrightarrow (\gamma-1){\prototypedist}_{i,j}^{-}(\x) + (1+\gamma){\prototypedist}_{i,j}^{+}(\x) \geq 0
\end{split}
\end{equation}
Next, we assume that the closest prototype with the correct label is fixed and denote it by $\vec{\prototype}_i$ -- we denote prototypes from the other class as $\vec{\prototype}_j$. In the end, we iterate over all possible closest prototypes $\vec{\prototype}_i$ and select the one that minimizes the objective (i.e. closeness to the original sample) -- note that this approximation drastically increases the number of optimization problems that must be solved and thus the overall complexity of the final algorithm. We then can rewrite~\refeq{eq:probcertainmeasure:constraint:convexapprox} as follows -- we make sure that~\refeq{eq:probcertainmeasure:constraint:convexapprox} is satisfied for every possible $\vec{\prototype}_{-}$:
\begin{equation}
\begin{split}
&(\gamma-1){\prototypedist}_{i,j}^{-}(\x) + (1+\gamma){\prototypedist}_{i,j}^{+}(\x) \geq 0 \\
&\Leftrightarrow (\gamma-1){\prototypedist}(\x,\vec{\prototype}_j) + (1+\gamma){\prototypedist}(\x,\vec{\prototype}_i) \geq 0 \quad \forall\,\vec{\prototype}_j\in \set{P}(\protolabel_{i})
\end{split}
\end{equation}
Applying even more simplifications yield:
\begin{equation}
\begin{split}
&(\gamma-1){\prototypedist}(\x,\vec{\prototype}_j) + (1+\gamma){\prototypedist}(\x,\vec{\prototype}_i) \geq 0\\
&\Leftrightarrow (\gamma-1)\left(\x^\top\distmat\x - 2\vec{\prototype}_j^\top\distmat\x + \vec{\prototype}_j^\top\distmat\vec{\prototype}_j\right) +\\&\quad (1+\gamma)\left(\x^\top\distmat\x - 2\vec{\prototype}_i^\top\distmat\x + \vec{\prototype}_i^\top\distmat\vec{\prototype}_i\right) \geq 0 \\
&\Leftrightarrow (\gamma-1)\x^\top\distmat\x + (1+\gamma)\x^\top\distmat\x - 2(\gamma-1)\vec{\prototype}_j^\top\distmat\x \\&\quad- 2 (1 + \gamma)\vec{\prototype}_i^\top\distmat\x + (\gamma - 1)\vec{\prototype}_j^\top\distmat\vec{\prototype}_j + (1 + \gamma)\vec{\prototype}_i^\top\distmat\vec{\prototype}_i \geq 0\\
&\Leftrightarrow 2\gamma\x^\top\distmat\x -2(\gamma - 1)\vec{\prototype}_j^\top\distmat\x - 2 (1 + \gamma)\vec{\prototype}_i^\top\distmat\x +\\&\quad (\gamma-1)\vec{\prototype}_j^\top\distmat\vec{\prototype}_j + (1+\gamma)\vec{\prototype}_i^\top\distmat\vec{\prototype}_i \geq 0\\
&\Leftrightarrow \x^\top\distmat\x + \q_j^\top\x + c_j \leq 0
\end{split}
\end{equation}
where
\begin{equation}
\begin{split}
\q_{j}^\top &= \frac{1}{-2\gamma}\left(-2(\gamma - 1)\vec{\prototype}_j^\top\distmat - 2 (1 + \gamma)\vec{\prototype}_i^\top\distmat\right)\\
c_{j} &= \frac{1}{-2\gamma}\left( (\gamma-1)\vec{\prototype}_j^\top\distmat\vec{\prototype}_j + (1+\gamma)\vec{\prototype}_i^\top\distmat\vec{\prototype}_i\right)
\end{split}
\end{equation}
Finally, we get the following convex quadratic optimization problem:
\begin{equation}\label{eq:probcertainmeasure:opt:convexapprox}
\begin{split}
&\underset{\xcf\,\in\,\RN^\dimsym}{\min}\;\pnorm{\xorig - \xcf}_1 \\
\text{s.t. }& \x^\top\distmat\x + \q_j^\top\x + c_j \leq 0 \quad \forall\,\vec{\prototype}_j\in \set{P}(\protolabel_{i})
\end{split}
\end{equation}
Note that we have to solve~\refeq{eq:probcertainmeasure:opt:convexapprox} for every possible closest prototype, every possible class different from the $i$-th class and finally for every possible class. Thus, we get the following number of optimization problems (quadratic in the number of classes):
\begin{equation}
|\setY| \cdot (|\setY|-1) \cdot P_N = |\setY|^2\cdot P_N - |\setY|\cdot P_N
\end{equation}
where $P_N$ denotes the number of prototypes per class used in the pair-wise classifiers.

Note that this number is much larger than $|\setY|$ which we got without introducing any surrogate or approximation. However, in contrast to~\refeq{eq:probcertainmeasure:opt}, the surrogate optimization problem~\refeq{eq:probcertainmeasure:opt:convexapprox} is much easier to solve because it is a convex quadratic program (convex QP) which are known to be solved very fast~\cite{Boyd2004}.

\section{Additional Empirical Results}\label{sec:appendix:experimens:goodnessexplanation}
\subsection{Probabilistic Reject Option}
The results for the different methods for computing counterfactual explanations of the probabilistic reject option~\refeq{eq:probareejct} are given in Table~\ref{appendix:table:experimentresults:rejectproba:blackbox}.
\begin{table}
\caption{Algorithmic properties -- Mean sparsity (incl. variance) of different counterfactuals, smaller values are better, for feasibility, larger values are better for feasibility.}
\centering
\footnotesize
\begin{tabular}{|c||c|c||c|c|c||}
 \hline
 \textit{DataSet} & Bb~\refeq{eq:cfreject:opt} Feasibility & Algo.~\ref{algo:cfreject:proba1} Feasibility & BlackBox~\refeq{eq:cfreject:opt} & Algo.~\ref{algo:cfreject:proba1} & Algo.~\ref{algo:cfreject:proba2} \\
 \hline
 Wine & $0.45 \pm 0.15$ & $0.45 \pm 0.21$ & $11.1 \pm 2.09$ & $13.0 \pm 0.0$ & $12.67 \pm 0.22$ \\
 Breast Cancer & $0.69 \pm 0.12$ & $0.49 \pm 0.02$ & $26.43 \pm 3.45$ & $30.0 \pm 0.0$ & $26.04 \pm 60.5$ \\
 t21 & $0.82 \pm 0.03$ & $0.4 \pm 0.24$ & $15.94 \pm 2.34$  & $17.39 \pm 0.64$ & $12.55 \pm 7.41$ \\
 Flip & $0.5 \pm 0.2$ & $0.4 \pm 0.24$ & $9.5 \pm 2.25$ & $11.75 \pm 0.23$ & $11.5 \pm 1.47$ \\
 \hline
\end{tabular}
\label{appendix:table:experimentresults:rejectproba:blackbox}
\end{table}

\end{document}